\documentclass[conference,a4paper]{IEEEtran}
\IEEEoverridecommandlockouts
\usepackage{cite}
\usepackage{amsmath,amssymb,amsfonts}
\usepackage{algorithmic}
\usepackage{graphicx}
\usepackage{textcomp}
\usepackage{xcolor}
\usepackage{tabu}
\usepackage{color}
\def\BibTeX{{\rm B\kern-.05em{\sc i\kern-.025em b}\kern-.08em
    T\kern-.1667em\lower.7ex\hbox{E}\kern-.125emX}}
\begin{document}

\title{Class Activation Map generation by Multiple Level Class Grouping and Orthogonal Constraint\\
}

\author{\IEEEauthorblockN{Kaixu Huang\textsuperscript{1}, Fanman Meng\textsuperscript{2},  Hongliang Li\textsuperscript{3}, Shuai Chen\textsuperscript{4}, Qingbo Wu\textsuperscript{5}, King N.Ngan\textsuperscript{6}, \\
\textit{School of Information and Communication Engineering}\\
\textit{University of Electronic Science and Technology of China }\\
\textit{Chengdu, China}\\
\textit{\{\textsuperscript{1}kaixu\_huang, \textsuperscript{4}s-chen\}@std.uestc.edu.cn, \{\textsuperscript{2}fmmeng, \textsuperscript{3}hlli, \textsuperscript{5}qbwu, \textsuperscript{6}knngan\}@uestc.edu.cn}}}

\maketitle

\begin{abstract}
Class activation map (CAM) highlights regions of classes based on classification network, which is widely used in weakly supervised tasks. However, it faces the problem that the class activation regions are usually small and local. Although several efforts paid to the second step (the CAM generation step) have partially enhanced the generation, we believe such problem is also caused by the first step (training step), because single classification model trained on the entire classes contains finite discriminate information that limits the object region extraction. To this end, this paper solves CAM generation by using multiple classification models. To form multiple classification networks that carry different discriminative information, we try to capture the semantic relationships between classes to form different semantic levels of classification models. Specifically, hierarchical clustering based on class relationships is used to form hierarchical clustering results, where the clustering levels are treated as semantic levels to form the classification models. Moreover, a new orthogonal module and a two-branch based CAM generation method are proposed to generate class regions that are orthogonal and complementary. We use the PASCAL VOC 2012 dataset to verify the proposed method. Experimental results show that our approach improves the CAM generation.
\end{abstract}


\section{Introduction}
Generating class activation map (CAM) for deep classification networks plays an important role in computer vision. It is used to provide object regions from image-level labels, and is widely used in many weakly supervised computer vision tasks, such as segmentation\cite{Revisiting,Panoptic,Instance,FickleNet,Multi-Evidence}, detection\cite{TS2C,Self-produced,Min-Entropy,Zigzag}, and recognition\cite{Attention-GAN,Discriminative}.

Since image-level labels are very rough, extracting CAM is a challenging task. The existing methods extract CAM by firstly establishing the classification model for a set of given classes, and then analyzing the classification model to find the regions that support the image to be a certain class. In the past few years, the second step gets a lot of attention, and several efficient CAM extraction models have been proposed. It proves that these models localize the key regions of the objects well. Meanwhile, the regions highlighted by these models are usually small part region rather than the complete objects, such as the ``Head'' of ``Person'', because in the classification model, the discriminative regions used to distinguish a certain class are usually small and local.

Several efforts have been done to overcome such drawback, and two strategies are proved to be useful. One is erasing strategy, which erases the regions already highlighted, and generates CAM again to highlight more regions. The other is to increase the perception area of each convolution operation by the dilated convolution, and thus highlights more regions. However, these strategies are based on single classification model, which contains finite discriminative cues that limit the CAM extraction. Note that CAM generation is sensitive to the classification models. By the manner of grouping classes into multiple class sets, multiple classification models can be established, which can provide more discriminative cues and highlight more object regions.

\begin{figure}[t]
\begin{center}
   \includegraphics[width=0.9\linewidth]{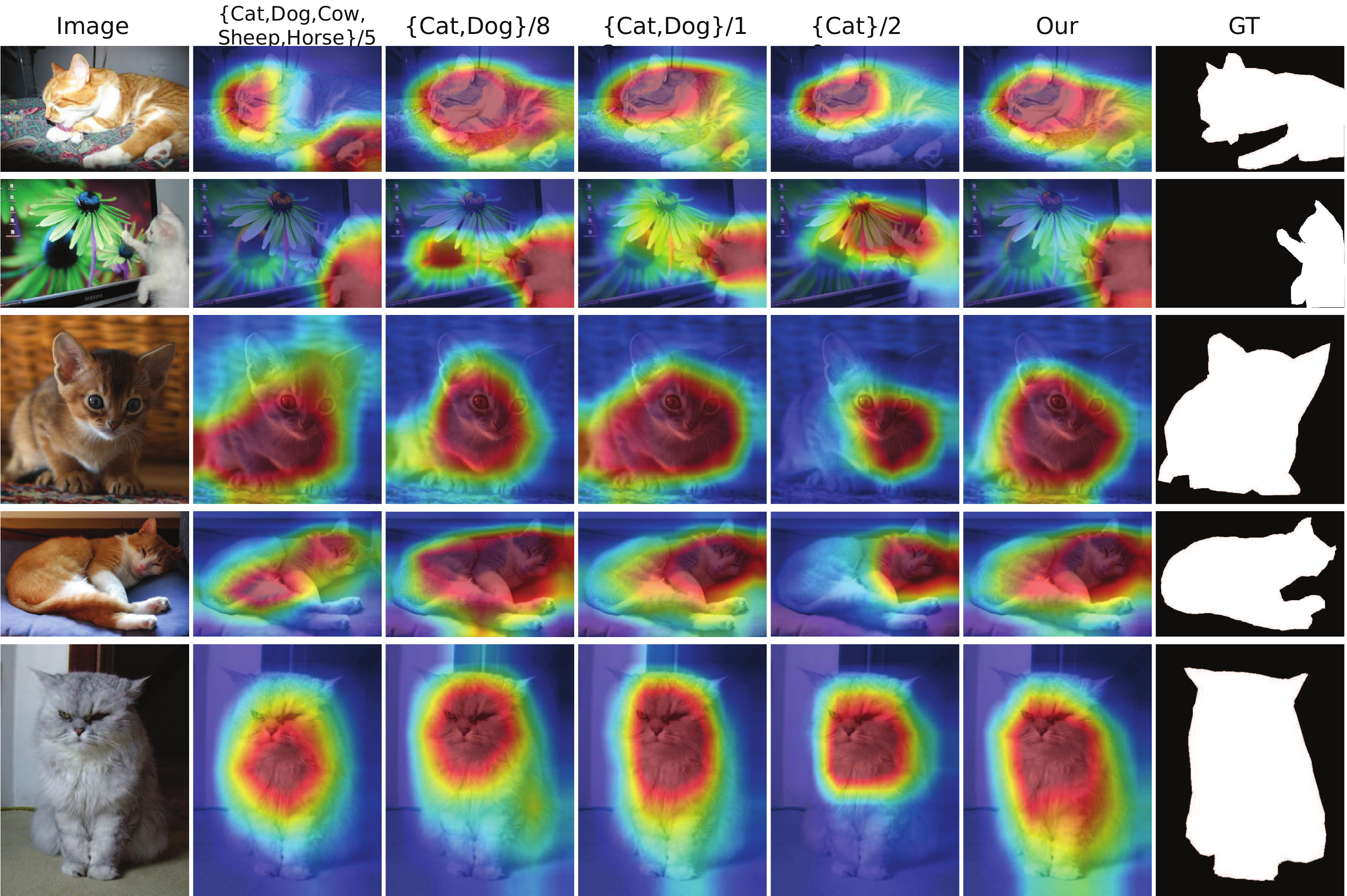}
\end{center}
   \caption{First column: initial images. Second - Fifth columns: the CAM results by different level of clustering. Sixth column: the combination results. Last column: ground truth. $\{\cdot\}/N$: a cluster $\cdot$ by clustering classes into $N$ classes. }\label{cat}
\end{figure}

In order to observe the sensitivity of the CAM to the classification models, we display some experimental results in Fig. \ref{cat}, where the 20 classes in the Pascal VOC dataset are merged based on their similarities to form a set of class groups, and each group is treated as a new class to train the classification network. We also implement the merging process iteratively to obtain multiple level of class groups, and represent them by $N$ indicating the number of clusters. $\{\cdot\}/N$ means a cluster $\cdot$ of clustering result by clustering classes into $N$ clusters. The results are displayed in the second column to the sixth column. It is seen that the existing method (``\{Cat\}/20'') highlights ``head'' regions of ``Cat'' only, while the group of ``\{Cat,Dog\}/8'' highlights both the ``head'' and ``body'' regions. By combining these CAMs further, more object regions (``Our'') can be highlighted. This motivates us to use multiple sets of class groups to enhance CAM generation.

Based on such motivation, this paper proposes a new CAM generation method, which firstly clusters initial classes into multiple level of class groups to generate diverse CAMs, and then combines these CAMs to generate more comprehensive class activation map. The proposed approach consists of four steps. The first step calculates the class relationships based on deep features of the classes. Then, we use the relationship matrix to cluster the classes into groups using a merging strategy, which clusters the similar classes each time to obtain multiple level of clusters. The second step uses the multiple level of clusters to train multiple classification models. In the third step, a new CAM generation method based on a two-branch based network and feature orthogonal module are proposed to generate multiple CAMs from the multiple classification models. In the final step, the multiple CAMs are combined to form the final CAM.

The proposed method has two advantages. Firstly, the discriminative cues among multiple level of clusters are captured more sufficiently, which can lead to better CAM generation. Secondly, the different image groups carry more discriminative cues that can highlight more key part regions of the object. Hence, the object can be described more accurately. We verify the proposed method on PASCAL VOC 2012 dataset. Experimental results demonstrate the effectiveness of the proposed method.


\section{The Proposed Method}\label{Method}
\subsection{Overview}
\begin{figure}[t]
\begin{center}
   \includegraphics[width=0.98\linewidth]{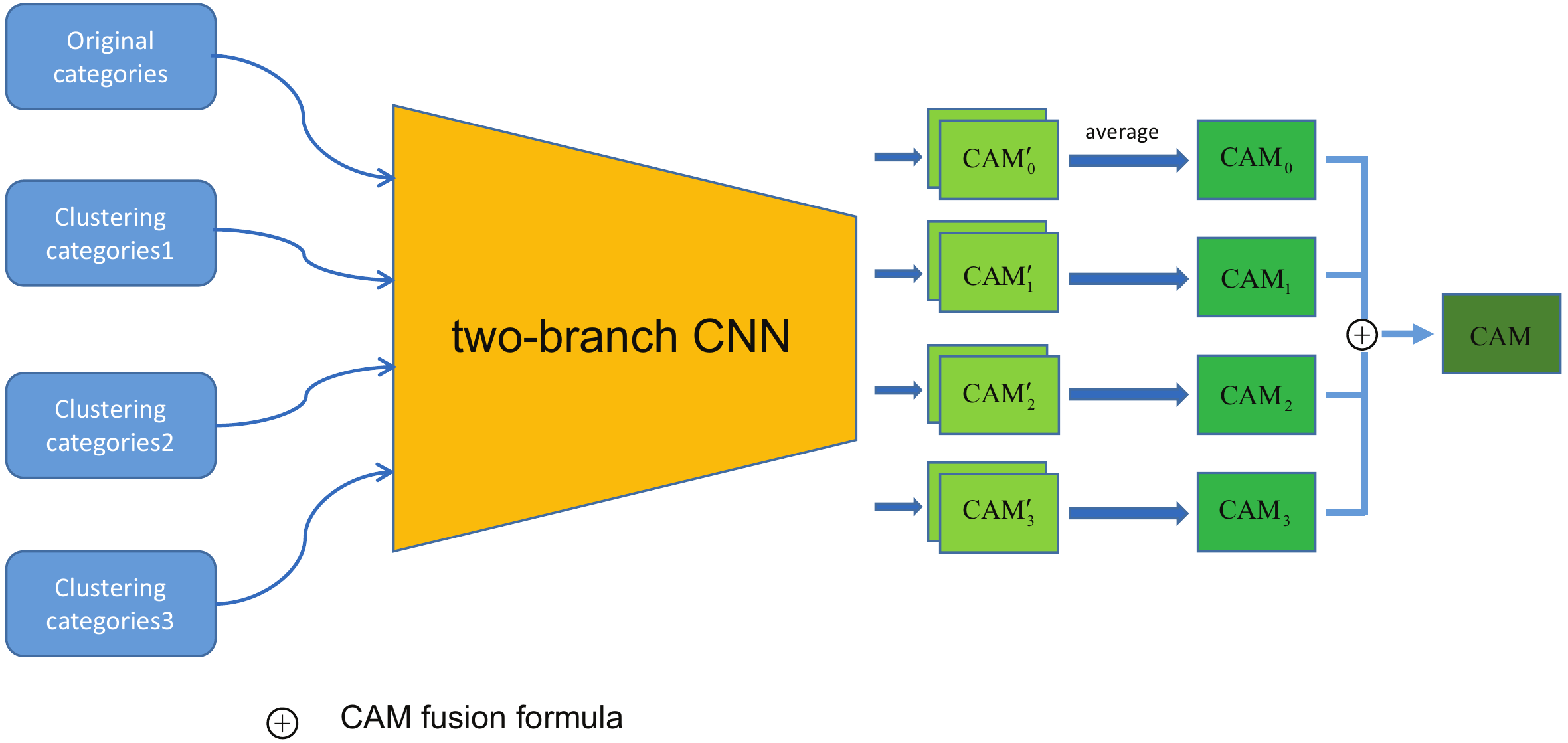}
\end{center}
   \caption{Overview of our approach. It consists of four steps such as the class clustering, classification network training, CAM generation and fusion. }\label{pipeline}
\end{figure}

Fig. \ref{pipeline} displays the flowchart of the proposed method, which consists of four steps such as the class clustering, classification network training, CAM generation and fusion.

In subsection \ref{Clustering}, the relationships between classes are calculated based on the deep features according to the classification network. Then, we use clustering strategy to merge the most similar classes and obtain clusters with different levels. In subsection \ref{Traning_orthogonal} and \ref{cam_branch}, in order to capture more discriminative information from the classification network, we propose a new two-branch-based CAM generation framework, which highlights different object regions by the two branches using feature orthogonal module. In subsection \ref{CAM_Generated}, the CAMs obtained from multiple classification models of different hierarchical clusters are combined to form the final CAM.

\begin{figure*}[t]
\begin{center}
   \includegraphics[width=0.98\linewidth]{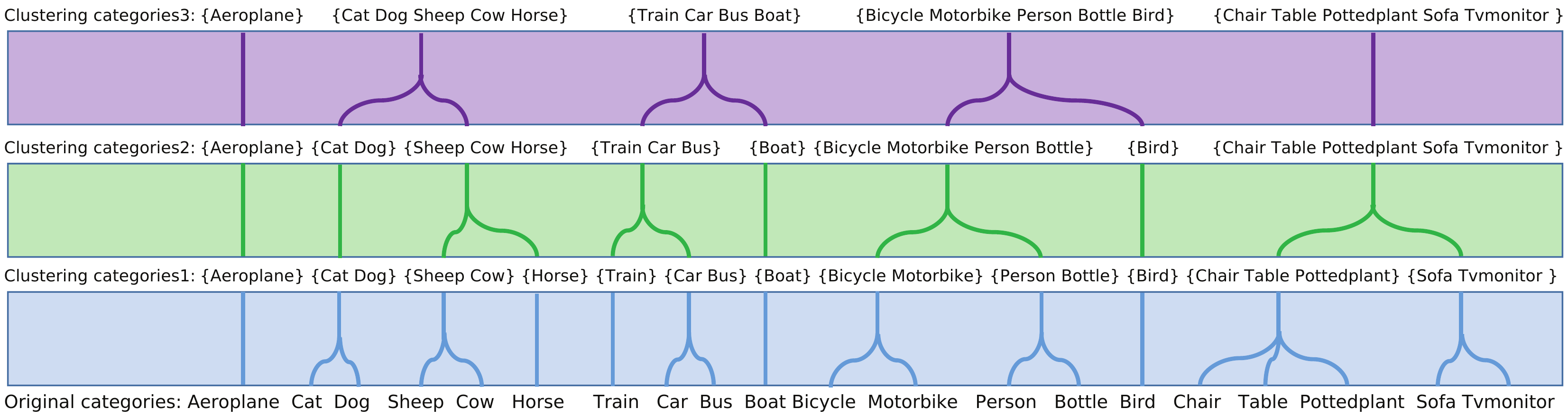}
\end{center}
   \caption{Blue, green, and purple represent the merging of categories for the first, second, and third level of clustering respectively. All categories in curly brackets are considered as a cluster. }\label{clustering_results}
\end{figure*}

\subsection{Category Clustering}\label{Clustering}
Single classification network trained by the entire classes is used in the existing CAM generation methods, which contains finite  discriminative cues to extract small regions. Here, we consider to merge classes into a set of class groups, and treat each class cluster as new class to obtain new classification network. Since the new classification network is totally different from the traditional classification network, different discriminative regions can be obtained. By implementing the merging process iteratively, more regions can be highlighted.

Note that clustering classes randomly is a simple way to accomplish such task, which however ignores the relationships between the categories that are important to CAM generation. Here, we implement class clustering based on the similarities between classes.

\begin{figure}[t]
\begin{center}
   \includegraphics[width=0.98\linewidth]{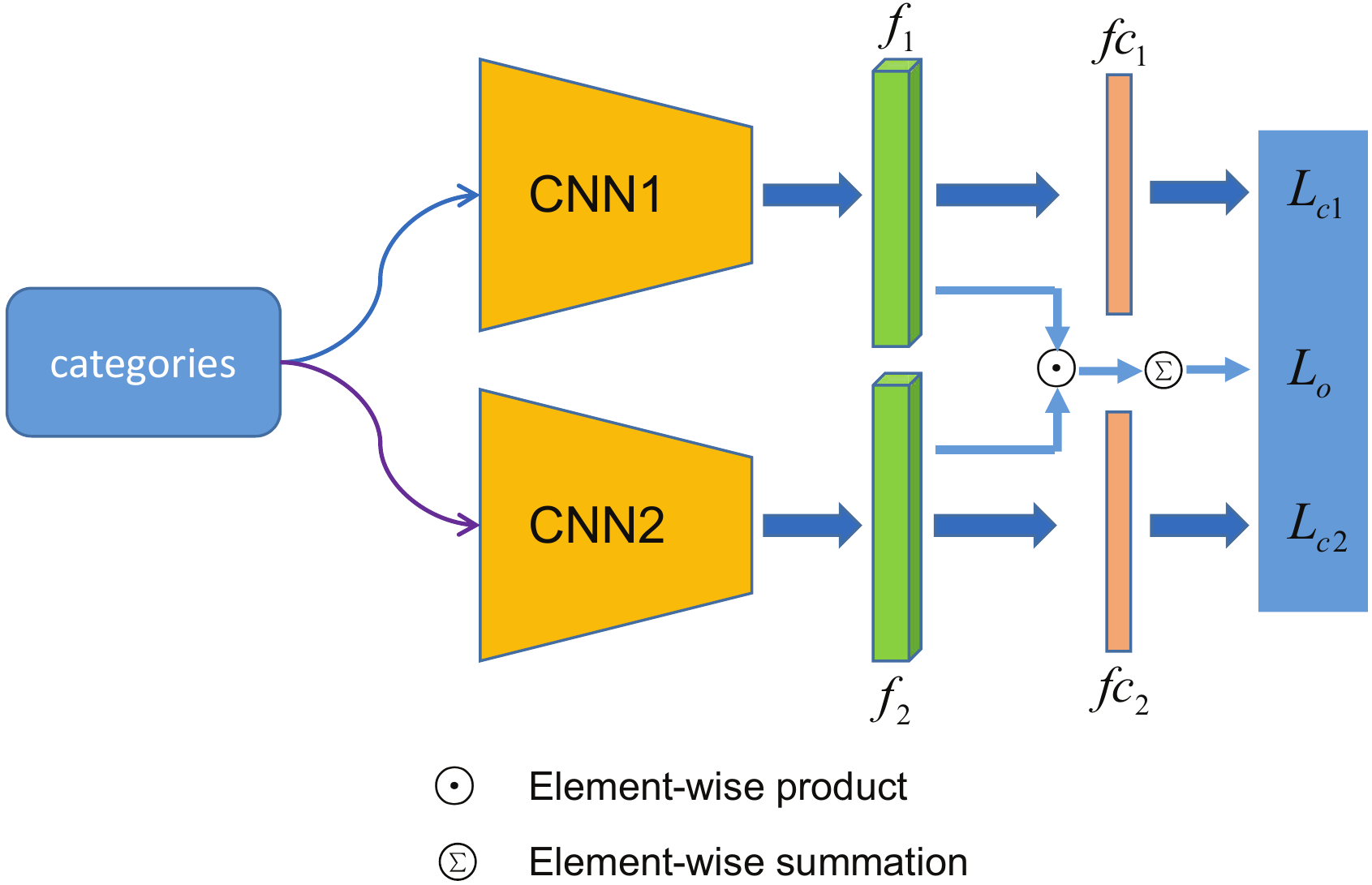}
\end{center}
   \caption{Detailed illustration of our CAM extraction model. It consists of two sub-branches that do not share parameters. We force the features of the two branches to be orthogonal by adding additional loss $L_o$. }\label{orthogonal}
\end{figure}

\subsubsection{Deep Feature for Each Class}\label{feature_information}
The key of calculating the similarity relationship is how to represent each class by a feature. To this end, we use deep classification network to represent the class. Specifically, we firstly train a classification network considering all classes. Then, the weight vector of the last FC layer that maps the convolutional feature to class labels is used as the feature. Let $n_c$ be the number of classes, the last FC layer contains $n_c$ nodes, of which the $i$th node corresponds to the $i$th class. We use the weight vector of the $i$th node as the feature of the $i$th class. It is seen that each class is described by a vector with length $n_k$, where $n_k$ is the number of nodes in the previous FC layer.

\subsubsection{Category feature clustering}
After representing each class $c_i$ by the feature vector $x_i$, we perform the clustering on these vectors by K-means algorithm to obtain class clusters $C_1$ with cluster number $N$, i.e.,
\begin{equation}
a_j=\frac{1}{n^c_{j}}\cdot \sum_{i\in C_j}x_i
\end{equation}
\begin{equation}
\arg\min(\sum_{j=1}^{N}\sum_{i\in C_j}{\left \| x_i-a_j \right \|_2})
\end{equation}
where $a_j$ represents the cluster center of the $j$th cluster, and $n^c_j$ represents the number of classes in the $j$th cluster. We set each class cluster as new class, and obtain the classification network $CNN_1$ for the first clustering.

Such clustering process is implemented based on $C_1$ again to obtain new clusters $C_2$.  Specifically, the clusters $C_k$ obtained by the $k$th level of clustering are used to train a new classification network, by which the cluster feature vectors can be obtained according to \ref{feature_information}. The $(k+1)$th level of clustering result $C_{k+1}$ is then obtained by K-means algorithm. Based on implementing such process iteratively, hierarchical clustering results are obtained. In Fig. \ref{clustering_results}, the final clustering results by implementing the clustering process three times are displayed, where different level of the class groups are used to form multiple level of classification networks.

\subsection{Training Classification Network with Feature-Orthogonal Module}\label{Traning_orthogonal}

Based on the clustering results, we next train the new classification model for the CAM generation. The traditional methods use single network for the classification, which leads to the insufficient extraction of the discriminative information. Here, we use two networks with different parameters for the classification. In addition, we use the orthogonal constraint on the deep convolutional features to force the two networks different, which therefore captures more complementary discriminative information.

The proposed two-branch based classification network is shown in Fig. \ref{orthogonal}, where two branches are used for classification, and feature orthogonal module that forces the descriptions of the two branches to be orthogonal is used to combine the two branches. Note that the orthogonal module has been used in the CAM generation to highlight different regions \cite{zhang2018adversarial}. However, it is based on the class activation map, i.e., making the regions orthogonal. Here, we formulate orthogonal module in the feature space, and make the deep feature orthogonal, which can force the two networks to capture diverse information while avoiding failure caused by insufficient deep descriptions. Specifically, we firstly design the classification model as a two-branch network, where the two network branches do not share parameters. Then, we propose feature orthogonal loss function to connect the two branches:
\begin{equation}
L_{o} = \left \|f_1\odot f_2\right \|_{sum}
\end{equation}
where $\odot$ represents the Hadamard product between the features $f_1$ and $f_2$, and $\left \|x \right \|_{sum} $ is the sum of all elements in $x$.
It is seen that the loss is large when the features are similar. Meanwhile, when the input features are dissimilar, the loss is small. Hence, such loss suppresses the output of the similar features and rewards dissimilar features. Therefore, complementary description can be obtained.
So far, we can define the overall loss function of the two-branch network as:
\begin{equation}
L_{all} = L_{c1}+L_{c2}+\lambda L_{o}
\end{equation}
where $L_{c1}$ and $L_{c2}$ represent the classical classification loss functions of the first and second branches, respectively. $\lambda$ is a hyper-parameter.

\subsection{CAM generation For Each Classification Network} \label{cam_branch}
\subsubsection{Extracting CAM from each branch}
For the input image, the CAM is firstly extracted from each classification branch. Here, we use the existing CAM generation method such as \cite{selvaraju2017grad} to obtain CAM of each branch simply.

\subsubsection{The CAM combination of two branches}
Two CAMs are obtained by the two branches. We next combine them by the average operation to obtain the final CAM, i.e. $M=\frac{1}{2} \sum_{i=1}^{2} M_{i}$.

\subsection{CAM Fusion}\label{CAM_Generated}
Based on the clustering method in Section \ref{Clustering}, hierarchical clustering results can be obtained, as shown in Fig. \ref{clustering_results}. Assuming the clustering algorithm is implemented $k$ times, a total of $k+1$ classification networks are obtained. Therefore, a total of $k+1$ CAMs are obtained.

We next combine these CAMs to obtain the final CAM. For an image, the CAM of a class such as ``Cat'' is obtained by
\begin{equation}
M=M_0+\frac{1}{k} \sum_{i=1}^{k} M_{i}-M'_0
\end{equation}
where $M_0$ is the CAM of the classes without considering clustering $C_0$, and $M_{i}$ is the CAM by the $i$th level clustering results for the class ``Cat''. $M'_0$ is CAM of the rest classes.

\section{Experiment}\label{Experiment}
\subsection{Experiment Setup}
\subsubsection{Dataset}
The PASCAL VOC 2012 \cite{pascal-voc-2012} is used for verification, which consists of 20 object categories. Training dataset with 10,582 images and validation dataset with 1449 images are both employed.

\subsubsection{Implementation Details}
In image normalization, the short sides of the images are resized to 224. Then, the center area with size $224\times 224$ is cut out from the image as the normalized image.

\begin{figure*}[h]
\begin{center}
   \includegraphics[width=0.8\linewidth]{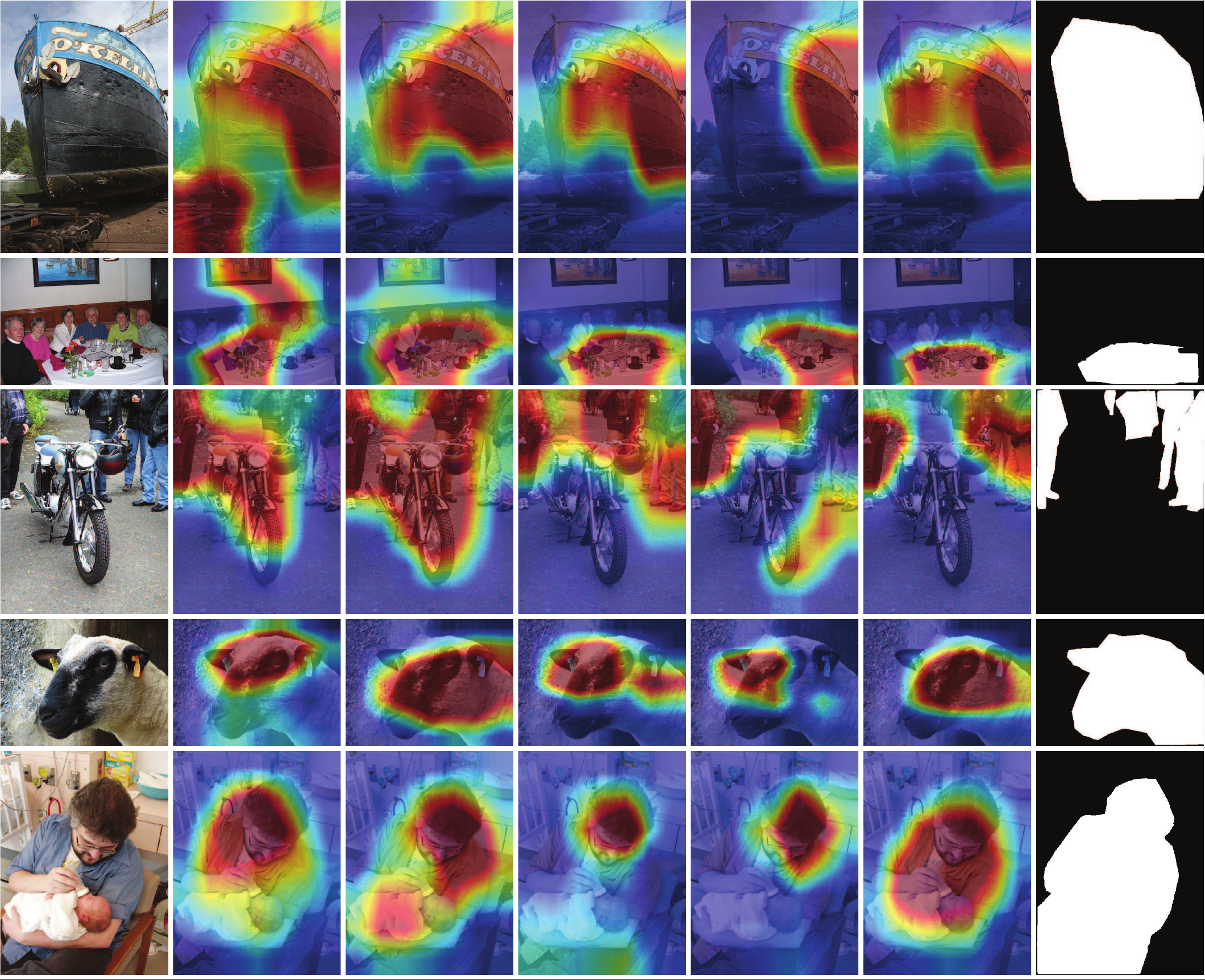}\\
   (a)~~~~~~~~~~~~~(b)~~~~~~~~~~~~~~(c)~~~~~~~~~~~~~(d)~~~~~~~~~~~~~~(e)~~~~~~~~~~~~~(f)~~~~~~~~~~~~~~(g)
\end{center}
   \caption{(a) Input image; (b)-(e) Results of clustering of 5, 8, 12, 20 respectively. (e) is also the result of baseline; (f) Results of our approach; (g) Ground truth. By comparing the results from Column (b) to Column (f), we can easily observe that our approach using relationship between categories greatly improves the results of CAM. }\label{results}
\end{figure*}

The classification network is initialized by the parameters pre-trained on the ImageNet. We train our network on the NVIDIA GeForce GTX1080 with 8GB memory and PyTorch 0.4 framework. We set the initial learning rate to 0.0001. When the decrease of loss within 5 epochs is smaller than a threshold, we reduce the learning rate by ratio 0.5. The batch size is set to 20. Each classification network is trained by 100 epochs.

In obtaining the feature vector of class for class clustering, we use ResNet-50\cite{he2016deep} as the baseline network to train the classification model. In order to balance the performance and training burden, we cluster the categories by four levels. The cluster numbers for the four levels are set to 20, 12, 8, and 5, respectively.

For the loss function, we set $L_{c1}$ and $L_{c2}$ as binary cross entropy loss, and set the hyper-parameter $\lambda=0.0001$ for $L_{o}$.

\subsubsection{Evaluation Criteria}
We use two widely used metrics to evaluate the performance of our approach and the comparison methods: mean intersection over union (mIoU) and mean localization errors values (mLEV). mIoU is the standard measure of semantic segmentation, which measures the fitness between CAM and ground truth (larger is better). Since CAM is a probability map rather than a binary mask, we use the threshold $ T = 0.15 $ which commonly used in CAM evaluation to binarize the class activation map. The mLEV indicates the ratio of the inaccuracy in the localization of the CAM (lower is better).

\begin{table}[t]
\small
\begin{center}
\caption{The mIoU values and mLEV values by the different clustering settings of our methods.}\label{clustering_times}
\begin{tabular}{ccccc}
\hline
Hierarchical Clustering Setting & mIoU & mLEV\\ \hline
Baseline N=\{20\}\cite{selvaraju2017grad} & 28.37/23.59 & 64.49/68.50\\ \hline
$N=\{20,12\}$ & 29.15/24.73 & 63.70/67.86\\ \hline
$N=\{20,12,8\}$ & 29.67/25.75 & 63.02/65.77\\ \hline
$N=\{20,12,8,5\}$& $\mathbf{30.24/25.97}$ & $\mathbf{61.82/65.08}$\\ \hline
$N=\{20,12,8,5,2\}$& 29.68/25.53 & 62.56/65.81\\ \hline
\end{tabular}
\end{center}
\end{table}

\begin{table}[t]
\small
\begin{center}
\caption{The mIoU values and mLEV values by the ablation studies of our methods.}\label{ablation}
\begin{tabular}{ccccc}
\hline
clustering & orthogonal & mIoU & mLEV\\ \hline
$\times$ & $\times$ & 28.37/23.59 & 64.49/68.50\\ \hline
$\times$ &\checkmark & 29.34/24.32 & 62.84/67.26\\ \hline
\checkmark  & $\times$ & 29.04/24.95 & 63.19/67.53\\ \hline
\checkmark  &\checkmark & $\mathbf{30.24/25.97}$ & $\mathbf{61.82/65.08}$\\ \hline
\end{tabular}
\end{center}
\end{table}

\subsection{Subjective Results}
The class activation maps of our approach are shown in the Fig. \ref{results}, where the original images, the CAMs of the multiple levels of clustering, the CAMs without clustering (i.e. the baseline method), and the CAMs by the proposed method are displayed. It can be seen that the activation maps of different clustering level are different. In addition, the class activation map of the proposed method is obviously better than the CAMs of different clustering levels and initial Grad-CAM, which proves the effectiveness of our approach on capturing more discriminative regions.

\subsection{Objective Results}

\subsubsection{The Results by Different Clustering Settings}
We first display the results of the proposed method by different clustering settings. The results are shown in Table  \ref{clustering_times}, where the number $N$ in the first column represents the clustering setting. For example, $N=\{20,12,8\}$ means a three-level clustering by clustering the classes into clusters with number of clusters 20, 12 and 8. The second column represents the mIoU of the validation set and the training set, and the third column represents the mLEV of the validation set and the training set.

As can be seen from Table \ref{clustering_times}, the mIoU values of the baseline (i.e. $N=\{20\}$) on the validation and training sets are 23.59\% and 28.37\%, and the mLEV values are 68.50\% and 64.49\%. When the clustering settings in our model are set to $\{20,12\}$, $\{20,12,8\}$, $\{20,12,8,5\}$, $\{20,12,8,5,2\}$, the mIoU on the validation dataset are 24.73\%, 25.75\%, 25.97\% and 25.53\%, and the mLEV are 67.86\%, 65.77\%, 65.08\% and 65.81\%, respectively. Moreover, mIoU values on the training set are 29.15\%, 29.67\%, 30.24\% and 29.68\%, respectively, and the mLEV are 63.70\%, 63.02\%, 61.82\% and 62.56\%. It can be seen the CAM generation is improved by using clustering strategy. Meanwhile, the results of $N=\{20,12,8,5\}$ are the best. This indicates that setting large number of levels is harmful to the CAM generation. The use of a certain number of clustering level is more useful to extract discriminative regions.


\begin{table}[t]
\small
\begin{center}
\caption{The mIoU values by the proposed method and the comparison methods.}\label{mIoU}
\begin{tabular}{ccccc}
\hline
Method&Network&PASCAL-val& PASCAL-train\\ \hline
CAM\cite{zhou2016learning} &ResNet-50 &20.90 & 24.69\\ \hline
Grad-CAM\cite{selvaraju2017grad} &ResNet-50&23.59 &28.37\\ \hline
CBAM\cite{woo2018cbam} &ResNet-50&19.93 & 23.85\\ \hline
Our & ResNet-50& $\mathbf{25.97}$ & $\mathbf{30.24}$\\ \hline
CAM\cite{zhou2016learning} & VGG-16 &23.81 & 27.35\\ \hline
Grad-CAM\cite{selvaraju2017grad} & VGG-16&27.62 &30.79\\ \hline
ACoL\cite{zhang2018adversarial} & VGG-16&19.52 & 20.41\\ \hline
Our & VGG-16&$\mathbf{31.32}$ & $\mathbf{34.37}$\\ \hline
\end{tabular}
\end{center}
\end{table}

\begin{table}[t]
\small
\begin{center}
\caption{The mLEV values by the proposed method and the comparison methods.}\label{mLEV}
\begin{tabular}{ccccc}
\hline
Method&Network&PASCAL-val& PASCAL-train\\ \hline
CAM\cite{zhou2016learning} &ResNet-50 &74.33 & 70.32\\ \hline
Grad-CAM\cite{selvaraju2017grad} &ResNet-50&68.50 &64.49\\ \hline
CBAM\cite{woo2018cbam} &ResNet-50&77.04 & 71.47\\ \hline
Our & ResNet-50& $\mathbf{65.08}$ & $\mathbf{61.82}$\\ \hline
CAM\cite{zhou2016learning} & VGG-16 &68.40 & 65.57\\ \hline
Grad-CAM\cite{selvaraju2017grad} & VGG-16&65.62 &63.24\\ \hline
ACoL\cite{zhang2018adversarial} & VGG-16&76.56 & 75.32\\ \hline
Our & VGG-16&$\mathbf{58.84}$ & $\mathbf{55.89}$\\ \hline
\end{tabular}
\end{center}
\end{table}

\subsubsection{Comparisons with Existing Methods}
We compare our approach to several existing methods of generating class activation map, such as CAM \cite{zhou2016learning}\footnote{https://github.com/metalbubble/CAM}, Grad-CAM \cite{selvaraju2017grad}, ACoL \cite{zhang2018adversarial}\footnote{https://github.com/xiaomengyc/ACoL} and CBAM \cite{woo2018cbam} \footnote{https://github.com/Jongchan/attention-module}. Since these methods do not give models based on the PASCAL VOC 2012 dataset, we use the code published by the author to training model based on the dataset. For CAM and Grad-CAM, we use ResNet-50 and VGG-16\cite{simonyan2014very} as backbone networks to generate CAM. For ACoL, we use the VGG-16 recommended in the code as the backbone. For CBAM, ResNet-50 is used (for fair comparison) as the backbone network.

The mIoU and mLEV values of the existing methods and our method are shown in Table \ref{mIoU} and Table \ref{mLEV}, where two types of network such as ResNet-50 and VGG-16 are employed as the backbone networks. It is seen from the tables that the proposed method is superior to the existing methods on both training and validation dataset, because our method can capture more discriminative cues by using multiple classification models, which results in the generation of more complementary and better class activation map.

\subsubsection{Ablation Study}
In this subsection, we conduct ablation study on our approach. Table \ref{ablation} shows the results of the ablation study by whether using the category clustering method and the feature orthogonal module or not. The manners of their combinations are shown in the first and second columns. The third to fourth columns represent the mIoU and mLEV values of the training set and the validation set. The larger the value of mIoU, the better the CAM extraction, and the smaller the value of mLEV, the more accurate the CAM localization.

The results of the ablation experiments of our proposed method are shown in Table \ref{ablation}. When we use the proposed feature orthogonal module only, mIoU values on the validation and training sets increase by 0.73\% and 0.97\% to the baseline method (without using both the clustering and orthogonal method), and mLEV on the validation and training sets decrease by 1.24\% and 1.65\% respectively. When we use the category clustering method only, the mIoU and mLEV values are both improved (1.36\% and 0.67\%, and 0.97\% and 1.30\% respectively). When we use both the category clustering method and the feature orthogonal module, mIoU and mLEV values are further improved (2.38\% and 1.87\%, and 3.42\% and 2.67\% respectively), which demonstrates the usefulness of the proposed method by using clustering strategy and orthogonal module simultaneously.

\section{Conclusion}
This paper proposes a new class activation map generation method, which extracts CAM by multiple classification networks. A hierarchical clustering method based on class relationships is firstly proposed to cluster classes into multiple level of clusters. Then, the clusters are treated as new classes to train multiple classification networks. To generate CAM more accurate, a new two-branch based network is proposed for training, and an orthogonal module is proposed to make the CAMs of the two branches different. Finally, the fusion method is proposed to combine the CAMs of the multiple networks, and generate the final CAM. The experimental results show that our method improves CAM generation in terms of larger mIoU values and smaller mLEV values.

\section*{Acknowledgment}
This work was supported in part by the National Natural Science Foundation of China under Grant 61871087, Grant 61502084, Grant 61831005, and Grant 61601102, and supported in part by Sichuan Science and Technology Program under Grant 2018JY0141.

\bibliographystyle{ieeetr}
\bibliography{sample-base}
\end{document}